\documentclass[...]{amsart}

\usepackage{graphicx}

\newtheorem{theorem}{Theorem}[section]

\newtheorem{definition}[theorem]{Definition}

\newtheorem{proposition}[theorem]{Proposition}

\begin{document}

\title{Typed Topological Structures Of Datasets }
\author{Wanjun Hu }
\maketitle

\begin{abstract}
A datatset $X$ on $R^2$ is a finite topological space. Current research of a dataset focuses on statistical methods and the algebraic topological method \cite{carlsson}. In \cite{hu}, the concept of typed topological space was introduced and showed to have the potential for studying finite topological spaces, such as a dataset. It is a new method from the general topology perspective. A typed topological space is a topological space whose open sets are assigned types. Topological concepts and methods can be redefined using open sets of certain types. In this article, we develop a special set of types and its related typed topology on a dataset $X$. Using it, we can investigate the inner structure of $X$. In particular, $R^2$ has a natural quotient space, in which $X$ is organized into tracks, and each track is split into components. Those components are in a order. Further, they can be represented by an integer sequence. Components crossing tracks form branches, and the relationship can be well represented by a type of pseudotree (called typed-II pseudotree). Such structures provide a platform for new algorithms for problems such as calculating convex hull, holes, clustering and anomaly detection.
\end{abstract}
\maketitle

\section{Introduction}

Machine learning algorithms, especially unsupervised algorithms, rely on the inner structures of datasets. For instance, the DBSCAN algorithm \cite{ester}, \cite{zhu1} needs to identify core points, then finds the reachable points within a radius and extends to a network. Outliers are detected afterward. The k-mean algorithm \cite{macqueen} randomly selects k-many points as centers of clustering.  Then it assigns each point to the designated cluster by calculating the distance and update the new center. The $k$-nearest neighbor algorithm \cite{cover} is another clustering algorithm that is based on the distance. The PCA algorithm 
\cite{pearson}, \cite{hotelling} \cite{hotelling2} calculates the eigenvectors that can approximate the variances of multi-dimensional vectors (features), and use them to form datasets of low-dimensional. The class of hierarchical clustering algorithms uses several methods to build a hierarchy of clusters. The self-organizing map algorithm \cite{kohonen1}, \cite{kohonen2} organizes high dimensional datasets into low dimensional ones by their similarities.  
  To detect the convex hull and holes in a dataset, there are algorithms such as Graham Scan algorithm\cite{graham}, Jarvis match algorithm \cite{jarvis}, and Chan's algorithm \cite{chan}. 
The alpha-shape algorithm \cite{herbert}, \cite{melkemi} generalizes the regular convex hull by adding an edge between points when they are on the boundary of a generalized disk of radius $1/\alpha$, for the given real number $\alpha$. An alpha-shape is just the regular convex hull when $\alpha=0$.  
    For images, there are also algorithms to study the shapes and holes, besides the convolution neural network. For instance, there are connected-component labeling algorithm \cite{rosenfeld}, and Conture tracing algorithms
 \cite{freeman},  \cite{suzuki}. Some algorithms may not work well with a dataset whose topological property is discrete.

Above mentioned problems and related algorithms are studied separately. They do not rely on a certain type of mathematical structure of a finite dataset. The topological data analysis  \cite{carlsson}, \cite{chambers}, \cite{chazal} treats a finite dataset as the vertices of a simplicial complex such as Vietoris-Rips complex or Cech compliex, and then calculate the homology groups and persistent homology. 

The direct study of the mathematical structure of a dataset is lacking. That type of structure is more of general topology nature. Unfortunately, a finite dataset is discrete, which is not interesting at all in the general topological sense. 
   In \cite{hu}, the concept of typed topology was introduced and used to study the inner structure of a dataset, such as tracks, clustering, branches. It was shown that the DBSCAN algorithm can be well-versed as a topological concept, i.e., $p-cluster$. 
  In this article, we continue the spirit of typed topology and investigate other types such as shapes, directions, and the related typed topological structures. As a result, we wish to present inner mathematical structures of a general dataset. We call them "typed topological structures". Those structures such as integer representation, type-II pseudotree become the framework for algorithms to solve various problems.

The article is arranged as follows. 
In section 2, we give a brief overview of the the major concepts in typed topological space, especially the concepts of types, typed closure, tracks of typed closures, type-connected components, branches. In section 3, we define types that are based on shapes and directions. 
  In section 4, we build the structures of a dataset, and study the boundaries on each tracks and the  calculations of convex hulls and holes in a dataset.  
   In section 5, we show how to map regular curves (from equations) into that structure. In section 6, we show how to represent a dataset as an integer sequence. In section 7,  we show how to represent connected components from each tracks as a type of pseudotree, called type-II pseudotree.

\section{Typed Topological Spaces}

\indent A topology on a set $X$ is a family of subsets of $X$, denoted $\mathcal T$, satisfying: (1) $\emptyset$, $X\in\mathcal T$; (2) $\cap\{U_i: i=1,2,...,n\}\in\mathcal T$ for any $U_1, ..., U_n\in\mathcal T$; (3) $\cup\{U: U\in\mathcal U\}\in\mathcal T$ for any subfamily $\mathcal U\subseteq\mathcal T$. 
   A dataset on the XY-plane is a finite topological space (metric space) with the usual  Euclidean metric. 

The following concepts were introduced in \cite{hu}. We include them here for quick reference and self completion, since that research is new.
   A typed topological space is a 5-tuple $(X, \mathcal T, P, \leq, \{\sigma_x: x\in X\})$, where $(X, \mathcal T)$ is a topological space, $(P, \leq)$ is a partially ordered set, and $\sigma_x: \mathcal T_x\rightarrow P$ is a partial function that satisfying $\sigma_x(U)\leq \sigma_x(V)$ whenever $U\subseteq V$. Here, $\mathcal T_x=\{U\in\mathcal T: x\in U\}$. When $\sigma_x(U)=p$, we say $U$ is of type $p$ and write it as $p\vdash U(x)$.  
    Given a type $p$ and a subset $A\subseteq X$, one can define the $p$-direct closure as $p\vdash CL_1(A)=\{y: A\cap U(y)\neq\emptyset~for~all~p\vdash U(y)\}$. Then $p\vdash CL_2(A)$ is defined as $p$-direct closure of $p\vdash CL_1(A)$, and so on. The $p\vdash tr(A)$ is defined as $A\cup \bigcup\{p\vdash CL_n(A): n=1,...\}$. 
   When the typed topological space has the so-called least $p$-neighborhood $U_{min}$ of $x$ for any $x\in X$ and any $p\in P$, denoted $p\vdash U_{min}(x)$, one has $p\vdash tr(p\vdash tr(A)) = p\vdash tr(A)$ holds for any $A\subseteq X$. That condition is easily met when the space $X$ is finite. For instance, 
 one can assume that if $\sigma_x(U)=p$ and $\sigma_x(V)=p$, then $\sigma_x(U\cap V)$ is defined and $\sigma_x(U\cap V)=p$. 
   The tracks of $p\vdash tr(A)$ are defined as: (1) $p\vdash Track_0(A)=A$; (2) $p\vdash Track_1(A) = p\vdash CL_1(A)\setminus A$; (3) for $n>1$, $p\vdash Track_n(A)=(p\vdash CL_n(A))\setminus (p\vdash CL_{n-1}(A))$.
       A subset $A\subseteq X$ is called type-p-connected, if there do not exist two collections of $p$-neighborhoods $\{p\vdash U(x_i): i\in I\}$ and $\{p\vdash U(x_j): j\in J\}$ satisfying: 
       (1) both $I\neq\emptyset$ and $J\neq\emptyset$; 
       (2) $A=\{x_i: i\in I\}\cup\{x_j: j\in J\}$; and 
       (3) $(\bigcup\{U(x_i): i\in I\})\cap (\bigcup\{U(x_j): j\in J\})=\emptyset$. When $X$ has the least $p$-neighborhood $U_{min}(x)$ of $x$ for any $x\in X$ and $p\in P$, an equivalent definition is that $A$ is type-p-connected if for every partition of $A$ with $A=I\cup J$, one must have $(\bigcup\{U_{min}(x_i): i\in I\})\cap (\bigcup\{U_{min}(x_j): j\in J\})\neq\emptyset$. 
    For any $A\subseteq X$ and a type $p$, each track $p\vdash Track_t(A)$ is the disjoint union of type-p-connected components. In other words,
      $p\vdash Track_t(A) = C^{t}_1\cup C^t_2\cup ... \cup C^t_{i_t}$ where $i_t$ is the number of type-p-connected components in the $t^{th}$ track $p\vdash Track_t(A)$.  
       A branch of $p\vdash tr(A)$ is defined as a sequence of $C^t_{j_i}$ where $t=t_0,t_0+1,t_0+2,..., t_0+m$ for some $t_0\geq 0$ and $j_i<i_t$ satisfying $C^{t+1}_{j_{i+1}}\cap (p\vdash CL_1(C^t_{j_i}))\neq\emptyset$.  These are the contents we adopt directly from \cite{hu}.

Above concepts can be generalized to a set of types.
In a typed topological space, for any subset $Q\subseteq P$ of types, 
and any subset $A\subseteq X$, one can define 
$Q\vdash CL_1(A)$, $Q\vdash CL_2(A)$, ..., and 
$Q\vdash tr(A)$ in the same way. For instance,
$Q\vdash CL_1(A)=\{y: ~\exists p\in Q, for~all~p\vdash U(y), A\cap U(y)\neq\emptyset\}$.

  The tracks of $Q\vdash tr(A)$ are defined likewise as below.
  \begin{enumerate}
  \item $Q\vdash Track_0(A) = A$;
  \item $Q\vdash Track_{t+1}(A) = (Q\vdash CL_{t+1}(A))\setminus (Q\vdash CL_t(A))$; and
  \item $Q\vdash Track(A) = \bigcup\{Q\vdash Track_t(A): t=0,1,2,...\}$
\end{enumerate}   

For finite space $X$, one will have $Q\vdash Track(A) = \bigcup\{Q\vdash Track_t(A): t=0,1,..., t_0\}$ for some non negative integer $t_0$. The smallest such $t_0$ is denoted $|Q\vdash Track(A)|$.

  The definition of type-Q-connectedness needs more scrutinization. In the DBSCAN algorithm, the connectedness was expressed by the so-called being "reachable". When a set of types $Q$, such as  lengths are involved, we want to be sure that
a point that is reachable in one distance $r$ in $Q$, then it should be reachable in the whole set $Q$. In our language, that can be translated into this,    
   "if a subset $A\subseteq X$ is type-p-connected for some $p\in Q$, then it is also type-Q-connected when $p\in Q$". 
      We can use the following definition.

\begin{definition}\label{type_Q_connect}
Let $(X, \mathcal T, P, \leq, \{\sigma_x: x\in X\})$ be a typed topological space. For any subset $Q\subset P$, a subset
$A\subseteq X$ is called type-Q connected if there exists a collection of typed open neighborhood 
 $\{p_x\vdash U(p_x, x): x\in A, p_x\in Q\}$ satisfying the following condition: (1) that collection covers $A$, i.e., $A\subseteq \bigcup\{p_x\vdash U(p_x, x): x\in A, p_x\in Q\}$; and (2) 
 for any partition $A=I\cup J$ with $I\neq\emptyset$ and $J\neq\emptyset$, and any choice of $U(p_x, x)$'s of $x$, 
 one has 
$(\cup\{p_x\vdash U(p_x, x): x\in I\})\cap (\cup\{p_x\vdash U(p_x, x): x\in J\})\neq \emptyset$.
\end{definition}    
  
This definition fits the definition of type-p-connected when the space is finite and has the least $p$-neighborhood property.

Similarly,   
   each track $Q\vdash Track_t(A)$ can be written as $C^t_1\cup C^t_2\cup ... \cup C^t_{i_t}$ for some integer $i_t>0$, where each $C^t_j$ is a type-Q-connected component, and $i_t$ is the number of type-Q-connected components in $Q\vdash Track_t(A)$. 
    A branch of $Q\vdash tr(A)$ of length $m$ is defined as a sequence 
    of $C^t_{j_t}$'s, i.e., $C^{t_0}_{j_0}, C^{t_0+1}_{j_1}, C^{t_0+2}_{j_2}$, 
    $..., C^{t_0+m}_{j_m}$, where $t_0\geq 0$, and for each $k=0,1,2,..., m, 
    j_k\leq i_{t_0+k}$. Further, the components satisfy
    $C^{t_0+k+1}_{j_{k+1}}\cap (p\vdash CL_1(C^{t_0+k}_{j_k}))\neq\emptyset$.

\section{Types Using Shapes And Directions}  

  A dataset on the XY-plane is a finite topological space, which is not interesting in general topology. When equipped with a set of types, it becomes interesting.  The merit about typed topological spaces is that we can choose any kind of types. For instance, we can choose types related to shapes, angles, directions, and even colors.
  
 In \cite{hu}, the left-r and right-r types are defined. They are types focus on left and right direction and a distance $r$, i.e., a $left$-$r$ neighborhood of a point $x=(a, b)$ is the set $\{y=(c,d)\in X: d(x,y)\leq r~and~ c\leq a\}$, and a $right$-$r$ neighborhood of $x=(a,b)$ is the set $\{y=(c,d)\in X: d(x,y)\leq r~and~ c\geq a\}$. Both neighborhoods are closed sets in the space $R^2$. However in the finite subspace $X$, they are actually open sets, according to the Theorem 5.4 in \cite{hu}.

In the following, we study sectors in a circle as types, a type that focuses on directions and distances. Such type provides good indication of data flow direction.

\begin{figure}
  \includegraphics[width=4in]{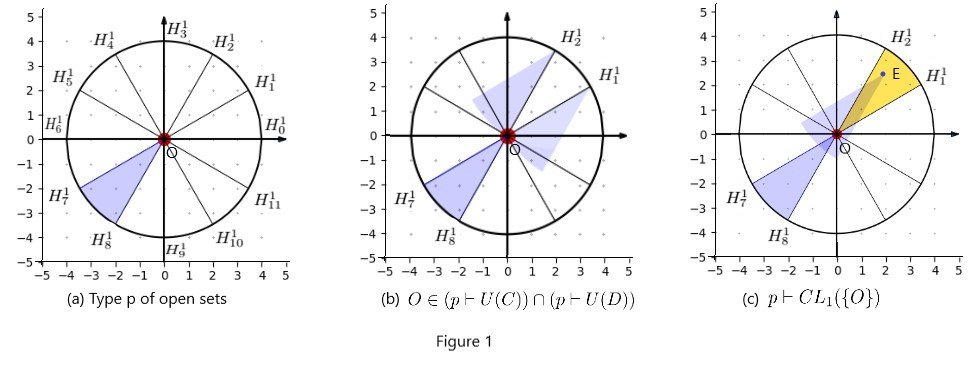}
  \caption{Sector As Type}
  \label{sector_as_type}
\end{figure}

   To do that, for a given positive real number $r>0$ and a given positive integer $n>0$, we divide a circle of radius $r$ into $n$ sectors of equal angle $\frac{360}{n}$. We enumerate each sector starting from the x-axis and counting counter-clockwise. In  Figure \ref{sector_as_type},  the radius of the circle is $r=4$. The center is at the origin $O=(0,0)$. The whole circle is divided into $n=12$ sectors with each of degree $\frac{360^o}{12} = 30^o$. Each of the sector indicates a direction from the origin. We put it in the following definition.
   
\begin{definition}\label{sectorType}
Let $n\geq 2$ be a natural number. For the circle with the center $x=(a,b)$ and radius $r$, we divide it into $n$ many sectors each of angle $360^o/n$. 
  The $i^{th}$ sector (i=0,1,...,n-1) starts from the vector $(a+r*cos(i*360^o/n), b+r * sin(i*360^o/n)$ and ends at the vector $(a+r*cos((i+1)*360^o/n), b+r * sin((i+1)*360^o/n)$. 
The $i^{th}$ sector is called the $i^{th}$ direction, and denoted $dir(r,i/(n-1))$ or just $dir(r, i)$ when $n$ is clear. $\Box$
\end{definition}

In Figure 1(a), the blue sector $H^1_7OH^1_8$ is the $7^{th}$ direction, or $dir(4, 7)$. For the point $H_1^2$ in Figure 1(b), the $dir(4, 7)$ neighborhood (i.e., $dir(4,7)\vdash U(H^1_2)$) is the blue area attached at $H^1_2$. Similarly, the blue area attached to the point $H^1_1$ is the $dir(4,7)$ neighborhood of $H^1_1$, i.e., $dir(4,7)\vdash U(H^1_2)$. In both cases, the point $O$ is inside the blue sector. Hence both points $H^1_2$ and $H^1_1$ are inside the direct closure of $\{O\}$, i.e.,  $\{H^1_1, H^1_2\}\subseteq dir(4,7)\vdash CL_1(\{O\})$.  

One can see that,for any two points $x,y$, their $dir(r, i)$ neighborhoods have parallel starting edges and parallel ending edges. In Figure 1(c), for any point $E$ inside the sector in Gold color, $O\in dir(4,7)\vdash U(E)$. Therefore, the following statement is true. 

\begin{proposition}\label{dir_closure}
Let everything be as in Definition \ref{sectorType}. Let $n=12$ and $x=O=(0,0)$ as in Figure \ref{sector_as_type}. Then $dir(4,7)\vdash CL_1(\{O\})$ is the sector in Gold color. $\Box$
\end{proposition}

Let $Q_r=\{dir(r, i): i=0,1,..., n-1\}$, i.e., the set of all $n$-many directions. Let also $P=\bigcup\{Q_r: r>0\}$. The set $P$ is the set of types for the typed topology $X$. 
For the origin $O$, the tracks, $Q_r\vdash Track_t\{O\}$ are subsets of the  concentric tracks in Figure 2 (a). Then, we can write $X=\{Q_r\vdash Track_t\{O\}: t=0,1,2,3,...\}$.
 To see that the tracks are the concentric disks of a dataset $X$, we may  
pad $X$ by another dataset. 
  For the number  $r$, set $X_{pad}=\{(a*r, b*r): a, b\in Z\}$, and let $X'=X_{pad}\cup X$. Then $X'$ forms a rather complete typed topological space in which $X$ is a subspace.  Inside $X'$,
  $Q_r\vdash Track_t\{O\}$ will be the $t^{th}$ concentric track as in Figure 2(a).

\begin{figure}
  \includegraphics[width=4in]{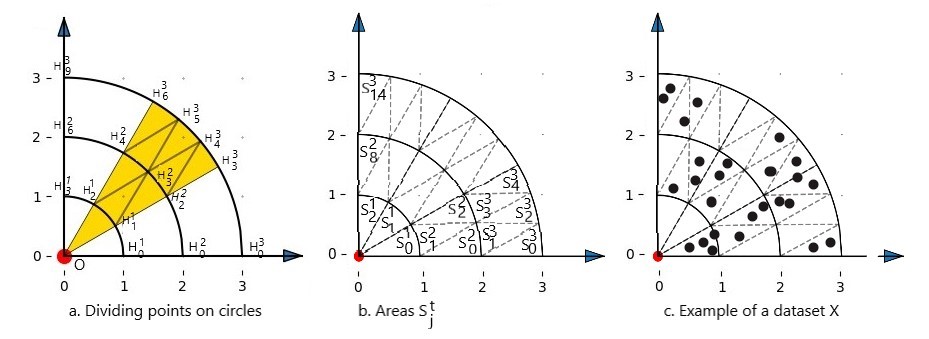}
  \caption{Tracks}
  \label{sector_tracks}
  \end{figure}

Each track can be further divided into small pieces. We first define some points. 
   
\begin{definition}\label{points_on_tracks}
Given a positive integer $n>1$, and a real number $r>0$, the sequence of dividing points $\{H^t_j: t>0, j=0,1,2,..., j<t*n\}$ is defined as follows:
\begin{enumerate}
\item For $t=0$, $H^0$ has only one point, i.e., $H^0_0=O$, the origin. 
\item For $t>0$, points in the set $H^t=\{H^t_k: k=0,1,..., tn\}$ are on the circle centered at the origin $O$ with radius $tr$,
and $H^t_k$ is identified by the angle $\theta^t_k= 360k/(tn)$.

\end{enumerate}
\end{definition} 

For $k=0,1,2,..., t*n-1$, we can write it as $k=i*t+j$ where $i<n$ and $j<t$. Here $i$ is the $i^{th}$-sector of the circle where $H^t_k$ locates, and $j$ is the $j^{th}$ point in the $i^{th}$ sector.

When $t=1$, the circle of radius $r$ is divided into $n$-many sectors. They are listed below as sequences of three points.
\\
\indent   $H^1_0, H^0_0, H^1_1, $\\
\indent   $H^1_1, H^0_0, H^1_2, $\\
\indent   $H^1_2, H^0_0, H^1_3, $\\
\indent $...,$\\
\indent   $H^1_j, H^0_0, H^1_{j+1}, $\\
\indent   $,...,$\\
\indent   $H^1_{n-1}, H^0_0, H^1_n$.

   When $t>1$, the $i^{th}$-sector ($i<n$), in $Q_r\vdash Track_t\{O\}$ is the $i^{th}$-sector of radius $tr$ minus the $i^{th}$-sector of radius $(t-1)r$. The corresponding arc of the sector is divided by $t$-many points $H^{t-1}_{(t-1)i},H^{t-1}_{(t-1)i+1}, ..., 
   H^{t-1}_{(t-1)i+t-1}$ in $H^{t-1}$ and the $t+1$-many points $H^t_{ti}, H^t_{ti+1},..., H^t_{ti+t}$ in $H^t$.
    We can list those points in the following sequences. 
\begin{align*}
     H^t_0, H^{t-1}_0, H^t_1, H^{t-1}_1, ..., H^t_{t-1}, H^{t-1}_{t-1}, H^t_t,& &  ... & & 0^{th}-sector\\
     H^t_t, H^{t-1}_{t-1}, H^t_{t+1}, H^{t-1}_{t}, ..., H^t_{t+t-1}, H^{t-1}_{t-1+t-1}, H^t_{t+t},& & ... & & 1^{th}-sector\\
     &... & &\\
	 H^t_{it}, H^{t-1}_{i(t-1)}, H^t_{it+1}, H^{t-1}_{i(t-1)+1},..., H^t_{it+t-1}, H^{t-1}_{i(t-1)+t-1}, H^t_{(i+1)t},& & ... & & i^{th}-sector\\
     & &... & &\\
     H^t_{(n-1)t}, H^{t-1}_{(n-1)(t-1)}, H^t_{(n-1)t+1}, H^{t-1}_{(n-1)(t-1)+1},...,  H^t_{nt},& &... & & (n-1)^{th}-sector.    
\end{align*}

  Each row in above sequence lists those points in  the $i^{th}$ block enclosed by points from $H^t$ and $H^{t-1}$. They are listed alternately starting with $H^t_{it}$ and ends with 
  $H^t_{(i+1)t}$. The corresponding start and end point in $H^{t-1}$ are $H^{t-1}_{i(t-1)}$ and $H^{t-1}_{i(t-1)+t-1}$, respectively. 
  Every three consecutive points in each row form a triangle like area, with a total of $2t-1$-many of them.     
    Those areas are marked as $S^{t}_{ti}, S^{t}_{ti+1}, ..., S^{t}_{ti+2t-2}$ as in Figure 2(b).  
   For $k=it+j<nt$, from $H^t_0$ to $H^t_k$, there are $i(2t-1)+2j$-many areas $S^t_0, ..., S^t_{i(2t-1)+2j-1}$. The set of all areas in $Q_r\vdash Track_t\{O\}$ is set as $\mathcal S^t=\{S^t_j: j=0,1,2,..., (2t-1)n\}$.

To make $S^t_k$'s pairwise disjoint, we can 
 set $S^t_k = S^t_k\setminus \bigcup\{S^t_j: j=0, 1, ..., k-1\}$. The new definition of $S^t_k$ removes 
the edge that is on border 
with $S^t_{k-1}$ for $k>0$.

Now, for any point $x=(a,b)\in R^2$, we can calculate the following information.

\begin{*}\label{algorithm1}
{\bf Algorithm 1.} Let $x=(a,b)\in R^2$. Compute the following integers.
\begin{enumerate}
\item the track number $t_x=ceiling (\sqrt{a^2+b^2}/r)$ such that $x\in Q_r\vdash Track_{t_x}\{O\}$,
\item the angle $\theta_x$ between the vector $\overrightarrow{Ox}$ (from origin to $x$) and $x-$axis, 
\item the smallest index $k_x<nt_x$ such that $\theta^{t_x}_{k_x}\geq \theta_x$, where $\theta^{t_x}_{k_x}$ is defined in Definition \ref{points_on_tracks}(2),
\item the sector index $i_x< n$ such that $k_x=i_xt_x+j_x$ for some $j_x<t_x$, which implies $x$ is in the ${i_x}^{th}$ sector of $Q_r\vdash_{t_x}\{O\}$,
\item the area index $w_x<(2t_x-1)n$ such that $x\in S^{t_x}_{w_x}$. 
\end{enumerate} 
\end{*}

The calculations for numbers in items 1-4 are straight-forward. For $w_x$ in item 5, we will have to judge among $S^{t_x}_{w-1}$, $S^{t_x}_w$, and $S^{t_x}_{w+1}$, where $w=(2t_x-1)i_x+2j_x$. 
When $j_x=0$, we will use the points 
  $H^{t_x}_{i_xt_x+j_x}, H^{t_x-1}_{i_x(t_x-1)+j_x}$,
  $H^{t_x}_{i_xt_x+j_x+1}, H^{t_x-1}_{i_x(t_x-1)+j_x+1}$. The area $S^{t_x}_w$ is formed by the three points   $H^{t_x}_{i_xt_x+j_x}, H^{t_x-1}_{i_x(t_x-1)+j_x}$, and 
  $H^{t_x}_{i_x t_x+j_x+1}$. The area $S^{t_x}_{w+1}$ is formed by the three points $H^{t_x-1}_{i_x (t_x-1)+j_x}$, 
  $H^{t_x}_{i_x t_x+j_x+1}, H^{t_x-1}_{i_x (t_x-1)+j_x+1}$.
  When $j_x>0$, we will use points
  $H^{t_x}_{i_xt_x+j_x-1}, H^{t_x-1}_{i_x(t_x-1)+j_x-1}$, 
  $H^{t_x}_{i_xt_x+j_x}, H^{t_x-1}_{i_x(t_x-1)+j_x}$,
  $H^{t_x}_{i_xt_x+j_x+1}, H^{t_x-1}_{i_x(t_x-1)+j_x+1}$. The area $S^{t_x}_{w-1}$ is formed by those points $H^{t_x}_{i_xt_x+j_x-1}$, $H^{t_x-1}_{i_x(t_x-1)+j_x-1}$, and 
  $H^{t_x}_{i_xt_x+j_x}$. The area $S^{t_x}_w$ and $S^{t_x}_{w+1}$ are the same as above. 
  To decide whether $x\in S^{t_x}_w$, we let 
  $\textbf{v}_1$ be the vector from $H^{t_x}_{i_xt_x+j_x}$ to $H^{t_x-1}_{i_x(t_x-1)+j_x}$, 
  $\textbf{v}_2$ be the vector from $H^{t_x}_{i_xt_x+j_x}$ to $H^{t_x-1}_{i_x(t_x-1)+j_x+1}$, and 
  $\textbf{v}_3$ be the vector from $H^{t_x}_{i_xt_x+j_x}$ to $x$. 
      Those three vectors share the same vertex $H^{t_x}_{i_xt_x+j_x}$, with  $\textbf{v}_1$ and $\textbf{v}_2$ being the two sides of $S^{t_x}_w$. If $x\in S^{t_x}_w$, then $x$ is between the two vectors $\textbf{v}_1$ and $\textbf{v}_2$. This can be checked by computing the three angles,   $\alpha$ as the angle between $\textbf{v}_1$ and $\textbf{v}_2$, $\alpha'$ as the angle between $\textbf{v}_1$ and
   $\textbf{v}_3$, and $\alpha''$ as the angle between $\textbf{v}_2$ and $\textbf{v}_3$. 
If both $\alpha'\leq \alpha$ and 
$\alpha''\leq \alpha$, then $x$ is between 
the two vectors $\textbf{v}_1$ and $\textbf{v}_2$. 
Instead of comparing $\alpha'$ with $\alpha$, we can directly calculate $cos(\alpha')$ and $cos(\alpha)$ by the dot product. 
Since the function $cos(x)$ is decreasing on $[0, 180^{\circ}]$, and $\alpha, \alpha', \alpha''$ are all between $0$ and $180^{\circ}$, 
to check whether $\alpha'\leq \alpha$, it is enough to check whether $cos(\alpha')\geq cos(\alpha)$. 
  Hence, 
   the point $x$ is in $S^{t_x}_w$ if both $cos(\alpha')\geq cos(\alpha)$ and $cos(\alpha'')\geq cos(\alpha)$ are true. The same method can be used to check whether  $x\in S^{t_x}_{w-1}$ or $x\in S^{t_x}_{w+1}$. The index $w_x$ is then decided accordingly.

\subsection{The quotient space $R^2/(r, n)$}

Based on above calculation, the space $R^2$ can be mapped to the family $\bigcup\{{\mathcal S}^t: t=1,2,3,...\}$, i.e., there is a unique function $f_S$ from $R^2$ to $\bigcup\{{\mathcal S}^t: t=1,2,3...\}$ such that $f_S(x) = S^{t_x}_{w_x}\in {\mathcal S}^{t_x}$. 
    Denote the family $\bigcup\{S^t: t=1,2,3...\}$ by 
$R^2/(r, n)$. 

  To define the metric in the quotient space $R^2/(r, n)$,  we first choose a representative point for each $S^t_j$. One way to do that is to use the centroid of the area $S^t_j$. However to facilitate other computation such as transformation that will be discussed below, we use a different point.
  Between the circles of radius $(t-1)r$ and $tr$, we consider the circle of radius $(t-1+t)r/2$. On that circle, we divide it into $(2t-1)n$-many arcs of equal lengths by points $\{s^{t}_j: j=0,1,..., (2t-1)n-1\}$, which are defined below.

\begin{enumerate}
\item $s^{t}_0$ is at the angle $360/(2(2t-1)n)$, which is roughly in the middle of $S^t_0$.
\item $s^{t}_j$ is at the angle $s^t_0+360j/((2t-1)n)$. 
\end{enumerate} 

Each $s^t_j$ is at the half angle position of either the arc $H^t_jH^t_{j+1}$ or the arc of the circle of radius $(t-1)r$ that connecting to $H^t_j$. For instance, in Figure 2 (a), $s^t_0$ is at the half angle position of the arc $H^2_0H^2_1$ and $s^t_2$ is at the half angle position of $H^2_1H^2_2$. The point $s^t_1$ is at the half angle position of $H^1_0H^1_1$.
   Certainly, we have $s^{t}_j\in S^{t}_j$. Hence, we can write $R^2/(r,n)$ as $\bigcup\{{\mathcal S}^t: t=1,2,3,...\}$ with ${\mathcal S}^t=\{s^t_j: j = 0,1,2,..., (2t-1)n\}$.

The distance $d$ between  $S^t_j$ and $S^{t'}_{j'}$ is then defined by the Euclidean distance between $s^t_j$ and $s^{t'}_{j'}$, i.e., $d(S^t_j, S^{t'}_{j'}) = d(s^t_j, s^{t'}_{j'})$. The set $R^2/(r, n)$ with the metric $d$ is a topological space. In a sense, it is the quotient space of $R^2$.

From $R^2$ to the quotient space $R^2/(r, n)$, the mapping of an equation $f(x,y) = 0$ or a function $y=f(x)$, can be defined using the following algorithm.

\begin{*}\label{algorithm2}
$\textbf{Algorithm 2. }$
Let $f(x,y)=0$ be an equation. For a sufficient large $t_0>0$ (which is so chosen that it serves as a window of the graph of $f$), 
 repeat the following for $t=1$ to $t_0$. \\
  (1)  Compute the intersection points of $f(x,y)=0$ and the circle of radius $tr$. Use above Algorithm 1 to calculate $f_S(x)$ for each intersection point $x$. Skip if no intersection point.\\
  (2)  For $k=0$ to $nt-1$,\\ 
  \indent the line from the origin $O$ to $H^t_k$ is $y=tan(\theta^t_k)x$;\\   
   \indent calculate the intersection points of the equation $f(x,y)=0$ and  $y=tan(\theta^t_k)x$;\\
\indent use above Algorithm 1 to calculate $f_S(x)=S^{t_x}_{w_x}$ for each intersection point $x$ if there exists any; otherwise skip.  
 \end{*}

\section{Branches, Boundaries, Holes}
A dataset $X\subseteq R^2$ may not have any meaningful order. However, the image $f_S(X)$ will have order structures. For $t>0$, define ${\mathcal S}^t(X):=\{S^t_j\in {\mathcal S}^t: f(x) = S^t_j~ for~some~x\in X\}$. It is the dataset $X$ restricted to $Q_r\vdash Track_t(\{O\})$. Further,  
$f_S(X) = \bigcup\{{\mathcal S}^t(X): t=1,2,3...\}$.

\subsection{Order structure on each track}
Each ${\mathcal S}^t(X)$ can be listed as $\{S^t_j: j\in B\subseteq \{0,1,2,..., (2t-1)*n\}$.
    According to the information in previous section,  
${\mathcal S}^t(X)$ can be writtn as $C^t_1\cup C^t_2\cup ...\cup C^t_{i_t}$ for some integer $i_t>0$, where each $C^t_l$ is the union of consecutive $S^t_j$'s, i.e., for some $0\leq j\leq j'\leq (2t-1)*n$, $C^t_l = \bigcup\{S^t_j\in {\mathcal S}^t(X): j\leq j\leq j'\}$. Hence, to calculate $C^t_1, C^t_2, ..., C^t_{i_t}$ inside ${\mathcal S}^t(X)$, we need to find maximum consecutive sequences. For some $0\leq j_1\leq j'_1< j_2\leq j'_2<j_3\leq j'_3<...<j_{i_t}\leq j'_{i_t}\leq (2t-1)*n$, we have

\begin{align*}
  {\mathcal S}^t(X) & & &\\
    = & \{S^t_j: j_1\leq j\leq j'_1\} & ... &\rightarrow C^t_1\\
    \cup & \{S^t_j: j_2\leq j\leq j'_2\} & ... &\rightarrow C^t_2\\
         & ... & ...\\
    \cup & \{S^t_j: j_{i_t}\leq j\leq j'_{i_t}\} & ...&\rightarrow  C^t_{i_t}\\
\end{align*}

Figure 2(c) shows an example of a dataset $X$.  Each black point is a point inside $X$. For that $X$, it is clearly that ${\mathcal S}^1(X)=\{S^1_0\}$. However for track 2, there seems to have two type-$Q_r$-connected components, $\{S^2_1, S^2_2\}$, $\{S^2_4, S^2_5, S^2_6, S^2_7\}$. However, the area $S^2_2$ and $S^2_4$ are only separated by one blank area. They are not far away enough. Similarly, in track 3, the area $S^3_0$ and $S^3_3$ are only separated by two blank areas, which is still too close. The situation between $S^3_7$ and $S^3_{12}$ is better, which are separated by 4 blank areas. So above arrangement of type-$Q_r$-connected components can be modified to combine two $C^t_l$'s if they are separated by less than 4 blank areas. Hence the actual type-$Q_r$-connected components of $X$ will be
$\mathcal S^1=\{S^1_0\}$, $\mathcal S^2 = \{S^2_1,S^2_2, S_2^4, S^2_5, S^2_6, S^2_7\}$, and $\mathcal S^3=\{S^3_0, S^3_3, S^3_4, S^3_5, S^3_6, S^3_7\}\cup\{S^3_{12}, S^3_{13}, S^3_{14}\}$. 

One major reason to choose 4 as the number of blank areas to separate two type-$Q_r$-connected components is that areas in the tracks $t$ and areas in track $t+2$ are separated by a distance of at least $r$. When considering a component $C^t_l$ and a component in $C^{t+1}_{l'}$ inside the same branch, we will measure the distance between them by the their vertices on the circle of radius $tr$. Two pieces of arcs, such as  $H^2_2H^2_3$ and $H^2_3H^2_4$ in Figure 2(a) allows up to 4 areas 
$S^3_5, S^3_6, S^3_7, S^3_8$ in track 3. Then, it is reasonable to say that $S^3_8$ and $S^3_7$ in track 3 is not inside the $Q\vdash CL_1(S^2_4)$. Hence they are separated and not inside the same branch. One special situation is when considering the four consecutive areas $S^3_2, S^3_3, S^3_4, S^3_5$, and the four consecutive areas $S^3_4, S^3_5, S^3_6, S^3_7$. Both have 4 areas. However on the circle of radius $2r$, they only span over one piece of arc $H^2_1H^2_2$ (resp. $H^2_2H^2_3$).

\subsection{Branches}

   As we mentioned in Section 2, a branch of $Q_r\vdash tr(\{O\}))$ is defined as a sequence 
    of $C^t_{j_t}$'s, i.e., $C^{t_0}_{j_0}, C^{t_0+1}_{j_1}, C^{t_0+2}_{j_2}$, 
    $..., C^{t_0+m}_{j_m}$, satisfying certain conditions.
   We can calculate the centroid of each $C^t_{j_t}$ and connect those centroids from one track to its next track. What we obtain is a simple graph of branches. 
Figure 3(b) shows two branches of the dataset for "d" which is the set of black points in Figure 3(a).

Each type-$Q_r$-connected component has the starting point and ending point on both circle of radius $tr$ (i.e., in $H^t$) and circle of radius $(t-1)r$ (resp. in $H^{t-1}$). Those on the circle of radius $tr$ are called $upper$ $starting$ $point$ (denoted $us$) and $upper$ $ending$ $poin$ (denoted $ue$), and those on the circle of radius $(t-1)r$ are called $lower~starting~point$ (denoted $ls$) and $lower~ending~point$ (denoted $le$). 
   For instance, assume that $C^t_{j_0}=S^t_0\cup S^t_1\cup...\cup S^t_5$ for a $t>0$. Then $us^t_{j_0} = H^{t}_0$,  $ue^t_{j_0}=H^{t}_3$,  $ls^t_{j_0}=H^{t-1}_0$,  and  $le^t_{j_0}=H^{t-1}_3$.

 Now for a branch $\mathcal B=\{C^{t_0+k}_{j_k}: k=0, ..., m\}$ for some $t_0$ and $j_k\leq (2(t_0+k)-1)n$, the set $LB=\{ls^{t_0+k}_{j_k}: k=0, 1,..., m\}\cup\{us^{t_0+k}_{j_k}: k=0,1,..., m\}$ will define the (lower) boundary of the branch. The 
 set $UB=\{le^{t_0+k}_{j_k}: k=0, 1,..., m\}\cup\{ue^{t_0+k}_{j_k}: k=0,1,..., m\}$ defines the (upper) boundary of the branch $\mathcal B$. 
  Figure 3(c) shows a  sequence of gold line segments that forms the lower boundary of the branch, and a sequence of green line segments that forms the upper boundary of the other branch. On each circle of radius $2r,  4r, 5r$, there are two points for an upper boundary and two points for a lower boundary. A hole is formed inside track $t=3$. The other two boundaries are not shown.

\begin{figure}
  \includegraphics[width=4in]{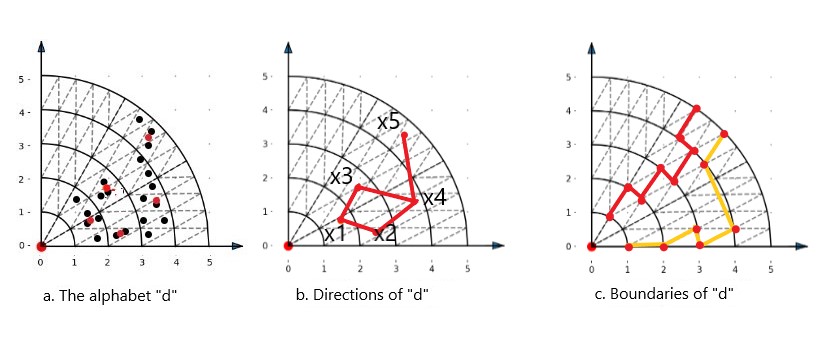}
  \caption{Directions And Boundaries}
  \label{sector_tracks}
  \end{figure}

Finding the convex hull of a dataset have been studied in literature \cite{jarvis}, \cite{graham}, \cite{herbert}, \cite{kirkpatrick}  
 For images, there are studies on finding holes,  \cite{rosenfeld}, 
 \cite{freeman},  \cite{suzuki}. 
     Our structure of upper and lower boundaries provides a new approach to find both convex hull and holes for any dataset. 
  The Graham Scan algorithm first finds the maximum and minimum $y$-coordinate for each $x$-coordinate in the dataset. Those form the top boundary and bottom boundary that wrap the whole dataset. Then one can start from the left most point and travel on 
both boundaries to decide whether a vertex should be kept as a vertex for the convex hull. It does backward checking to find the right vertex in previous checked vertices. It uses the strategy that is similar to binary search algorithm. Hence the time complexity is $n*log~n$. Our structure provides top boundary and bottom boundary using vertices that may not be inside $X$, but are close to vertices in $X$ by a distance $\leq r$. 
  
 Our structure also helps finding holes in any type of dataset $X$ including images. 
Let  
 $\mathcal B_1=\{C^{t_1+k}_{j_k}: k=0,1,..., m\}$
   and $\mathcal B_2$
be two branches. If for some $0\leq k_1< k_2\leq m$, both $C^{t_1+k_1}_{j_{k_1}}$ and  $C^{t_1+k_2}_{j_{k_2}}$ are also inside $\mathcal B_2$, and no $C^{t_1+l}_{j_l}$ is in $ {\mathcal B_2}$ for any $k_1<l<k_2$,
  then a $hole$ exists between the two branches.
The boundaries of the hole come from the boundaries of $\mathcal B_1$ and $\mathcal B_2$. 

An anomaly is a $C^t_j$ that is not inside any branch, which is automatically detected.

\subsection{Lines, Angles, Polygons By A  Dataset}

\indent In addition to convex hull or holes, we may also study the shape of $X$.

Inside $R^2$, a set of points $X=\{(a_j, b_j): j=0,1,...,m\}$ are on the same line $y=mx+b$ for $m=\frac{b_1-b_0}{a_1-a_0}$ and $b=b_1-m*a_1$, if and only if $ \frac{b_{j+1}-b_j}{a_{j+1}-a_j} = m$, or equivalently, if the angle between the vector 
$\overrightarrow{x_jx_{j+1}}$  and the $x-axis$ is the same as that of $\overrightarrow{x_0x_1}$ and $x-axis$. 

 A set
 $X=\{x_j: j=0,1,..., m\}$ in the quotient space $R^2/(r,n)=\bigcup\{\{s^{t}_j: j=0,1,2, ...,tn\}: t=1,2,3,...\}$
 may not form a straight line in the sense of a line in $R^2$. 
  A line in $R^2/(r, n)$ will have slightly different meaning. It will be a line but with some flexibility. The flexibility can be measured by the types. 
  
  Given any positive integer $n'>1$, for $x_j$ and $x_{j+1}$ inside $X$, we can use Algorithm 1 to calculate  $r'_j=distance (x_j, x_{j+1})$, and $i_j<n'$ the sector index of $x_{j+1}$ in the circle of $r'_j$ with center $x_j$, which is partitioned into $n'$ sectors. Then we have $x_{j+1}\in (dir(r'_j, i_j/(n'-1))\vdash CL_1(x_j))$. Then, the set $X$ is associated with the set of types $P_{n'}(Z)=\{dir(r'_j, i_j/(n'-1)): j=0,1,2,..., m-1\}$. 
   We say that a set $X$  forms a type $dir(*, i_0/(n'-1))$ line in $R^2/(r,n)$ if all $i_j=i_0$. The set $X$ forms a straight line in $R^2$ if, for any $n'>1$, it is a type $dir(*, i_0/(n'-1))$ line in $R^2/(r,n)$ for some $i_0$.    
   Put together, above argument can be stated as the following theorem.
 
 \begin{theorem}\label{s-line}
Given a set $X=\{s^{t_k}_{j_k}: k=0,1,2,...,m\}\subseteq R^2/(r,n)$,  
 it forms a straight line in $R^2$ if and only if it is true that, for any $n'>1$, $X$ is  a type $dir(*, i_0/(n'-1))$ line in $R^2/(r, n)$ for some $i_0<n'$.
$\Box$
\end{theorem}  

Further, we can define angles and edges of polygons in $R^2/(r,n)$.
Given $n'>1$,
let  $X=X_1\cup X_2$, where
$X_1=\{s^{t_k}_{j_k}: k=0,1,..., k_0-1\}$ and 
$X_2=\{s^{t_k}_{j_k}: k=k_0, ..., m\}$, for some $0<k_0<m$. If $X_1$ is a type $dir(., i_1/(n'-1))$ line and $X_2$ 
 is a type $dir(.,i_2/(n'-1))$ 
line with $i_1\neq i_2$, then $X$ forms a type $dir(., ?/(n'-1))$ angle with the vertex  at 
$s^{t_{k_0}}_{j_{k_0}}$.   
   If for some $0<k_0<k_1<...k_u<m$ such that 
$X_0=\{s^{t_k}_{j_k}: 0\leq k< k_0\}$ 
is a type $dir(., i_0/(n'-1))$ line, 
$X_1=\{s^{t_k}_{j_k}: k_0\leq k<k_1\}$ 
is type $dir(., i_1/(n'-1))$ line, ...,with $i_j\neq i_{j+1}$, then $X$ forms 
the $u$-sides of a polygon. A polygon of $u+u'$-sides will be formed by $X'$ and $X''$, where (1) $X'$ and $X''$ have the same starting point $s^{t_0}_{j_0}$ and ending point $s^{t_m}_{j_m}$; (2) $X'$ is $u$-sides;  and (3) $X''$ is $u'$-sides.

\section{Transformations in $R^2/(r,n)$}

In the quotient space $R^2/(r,n)$, a transformation is a mapping $f: \bigcup\{{\mathcal S}^t: t=1,2,...\}\rightarrow \bigcup\{{\mathcal S}^t: t=1,2,...\}$. 
One  simple case is a type of transformations $f(x)$ that satisfy $f({\mathcal S}^t)\subseteq {\mathcal S}^t$ for all $t=1,2,...$. Such a transformation can be decomposed into $f=\Sigma_{t=1}^\infty f_t$, 
where $f_t: {\mathcal S}^t\rightarrow {\mathcal S}^t$ and 
$f_t= f|_{{\mathcal S}^t}$. Each $f_t$ can be identified by a function 
from $\{0,1,2,..., (2t-1)n-1\}$ to $\{0,1,2,...\}$ and then $mod ((2t-1)n)$.

A $rotation$ of angle $\theta$ degree is such a transformation, 
 $\Sigma_{t=1}^\infty \{R^t_{\theta}: {\mathcal S}^t\rightarrow{\mathcal S}^t\}$ where 
$R^t_{\theta}: \{0,1,2..., (2t-1)*n-1\}\rightarrow \{0,1,2..., (2t-1)*n-1\}$ satisfies 
$R^t_{\theta}(j) = j+floor(\theta/\theta^t)$. 
 The angle $\theta^t$ is defined as  $\theta^t =  360/((2t-1)n)$, which is also
 the angle between two consecutive points $s^t_j$ and $s^t_{j+1}$. 
 Figure 4 (b) shows the result of rotating the dataset in Figure 4(a) by $\theta^1=360^{o}/n$ degree counter-clockwise.  
   It is easy to show that such a rotation is invertible. When $R^t_{\theta}(s^t_j) = s^t_{j'}$ after rotating $\theta$ degree, we also have $R^t_{-\theta}(s^t_{j'}) = s^t_j$. 

 \begin{figure}
  \includegraphics[width=4in]{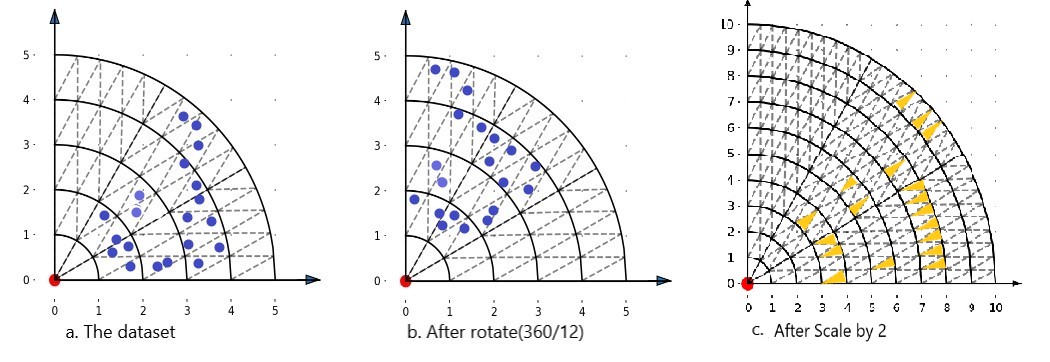}
  \caption{Transformations}
  \label{transformation}
  \end{figure}

Translation transformation is complicated. In general, a translation that translates origin to a point $(a,b)$ is a straight forward mapping. However, such a mapping may not work well in $R^2/(r,n)$. There will have clear distortion of any shape. Further, the inverse translation may not exist.  

    In the following, we provide a tailored version of translation, which is a translation along a line that passing origin. The ray from origin to a point $H^1_{i}$ with $0\leq i<n$ on the circle of radius $r$ is denoted $Ray_{i}$. 
$Ray_{0}$ is the $x$-axis. A translation of $k$ units along $Ray_{i}$, denoted $T_k$, is defined below.
\begin{align*}
T_k: R^2/(r,n)&\rightarrow R^2/(r,n)\\
    S^t_j &\rightarrow S^{t+k}_{j'}, j'=(2(t+k)-1)i+(j-(2t-1)i)
\end{align*}

An area $S^t_j$ in track $t$ is $j-(2t-1)i$ units away from the $Ray_{i}$. After the translation $T_k$, it is mapped to an area that is still $j-(2t-1)i$ units away from the $Ray_{i}$. With track $t+k$, the $Ray_{i}$ intersects at the area $S^{t+k}_{(2(t+k)-1)i}$. Hence $T_k$ is defined as above.
  The translation $T_k$ is still problematic. If one translates a dataset  $k>0$ units along $Ray_{i}$, and then translates $-k$ units backward along the same ray, it will return to  the same dataset. However, if one translates $k<0$ units first, and then translates back $-k$-units, it may not return to the same dataset.

   There are other ways to define translations. For instance, one can define a translation along $Ray_{i}$ that maps an area $S^t_j$ in track $t$ to an area in track $S^{t+k}$, while keeping some type of ratios. We skip those discussions.

 A scaling transformation of a factor $k>0$ can be defined as the mapping
\begin{align*}
S_k: R^2/(r,n)&\rightarrow R^2/(r,n)\\
    S^t_j &\rightarrow S^{ceiling(tk)}_{ceiling(jk)}.
\end{align*}
 It maps areas in track $t$ to areas in track $ceiling(tk)$, and multiply the index $j$ by $k$. Figure 4(c) shows an example of scaling the dataset in Figure 4(a) by a factor of $k=2$. Th areas in gold color shows the result. 
 
 In general, the scale transformation defined above may not map areas along a $Ray_i$ to areas still along $Ray_i$. For instance, in Figure 3(c), the area of $S^4_6$ in Figure 4(a) is mapped to $S^8_{12}$, which is not along the $Ray_1$. It may not map areas within the same sector. For example, the area $S^4_7$ in the $1^{th}$ sector is mapped to $S^8_{14}$ in the $0^{th}$ sector. Further it satisfies a similar  inverse property as the translation defined above, i.e., $S_{1/k}\circ S_{k} = identity$ is true only when $k$ is a positive integer. 

Similarly, one can define other scaling transformations. We will not discuss it here.

\section{Integer Sequence Representation}

The quotation space $R^2/(r,n)$ can be enumerated by $N$,  the non-negative integers. A quick observation is that there are $t^2n$ many areas in the first $t$-many tracks, and there are $t^2n-(t-1)^2n = (2t-1)n$ many elements in the $t^{th}$ track. The enumeration is arranged as follows, where the $t^{th}$ track is enumerated from $t^2n$ to $(t+1)^2n-1$.

\begin{enumerate}
\item  $\mathcal S^1=\{S^1_j: j=0,1,2,..., n-1\}$ has $n$-many elements. It is enumerated as $0, 1, ..., n-1=1^2n-1$. 
\item $\mathcal S^2=\{S^2_j: j=0,1,2,..., (2*2-1)n-1\}$ has $3n$-many elements. It is enumerated as $n, n+1, ..., 4n-1=2^2n-1$. 
\item $\mathcal S^t=\{S^t_j: j=0,1,2,..., (2t-1)n-1\}$ has $(2t-1)n$-many elements. It is 
enumerated as $(t-1)^2n, (t-1)^2n+1, ..., (t-1)^2n+(2t-1)*n-1=t^2n-1$.
\end{enumerate}

With above integer representation of areas , each dataset $X$ is represented as an integer sequence, also denoted $X$.
   With $n=12$, the dataset in Figure 2(c) is represent as {0, 13, 14, 16, 17, 18, 19, 48, 51,52,53,54,55,56,60,61, 62}. Similarly, the dataset in figure 3(a) and also in figure 4(a) are all represent by  {12, 14, 15, 17, 49, 54,55, 109, 110, 111, 112, 113, 114, 115, 117, 204, 205, 207}. The dataset in figure 4(b) is represent as    {15, 17, 18, 20, 54, 59, 60, 116, 117, 118, 119, 120, 121, 122, 124, 213, 214, 216}.
   
  When dividing each number $n_j$ in the sequence $X$ by $n$, $n_j=q_jn+r_j$, we obtain the $q_j$ and $r_j$. The will be a complete square, i.e., $q_j=t_j^2$, where $t_j$ is the track number where $n_j$ is. The number $r_j$ is the index of the corresponding area of $n_j$ inside the $t_j$-th track.  Using the dataset in figure 4(a) as an example, when dividing each number in the sequence by $n=12$ and taking the quotient, we obtain a  sequence, 
   $\{1,1,1,1, 4,4,4, 9,9,9,9,9,9,$ $9,9,16,16, 16\}$. We call it the $track$ $sequence$. The track sequence can be rewritten as 
   $\{1^2,1^2,1^2,1^2,$ $ 2^2,2^2,2^2, 3^2,3^2,3^2,3^2,3^2,$ $3^2,3^2,3^2,4^2, 4^2, 4^2\}$. It indicates that the dataset has 4-many tracks, i.e., 
\begin{enumerate}
\item the subsequence $\{12, 14, 15, 17\}$ is in the 2nd track within the circle of radius $(1+1)r=2*r$,
\item the subsequence $\{49, 55, 55\}$ is in the 3rd track within the circle of radius $(2+1)r=3*r$,
 
\item the subsequence  $\{109, 110, 111, 112, 113, 114, 115, 117\}$ is in the 4th track within the circle of radius $(3+1)r=4*r$, and 

\item the subsequence $\{204, 205, 207\}$ is in the 5th track within the circle of radius $(4+1)r=5r$.
\end{enumerate}   
    
    Integer representations of dataset under transformation have interesting properties.     
    For instance, after rotating the dataset in figure 4(a) by $360/12=30$ degree, the new dataset is in figure 4(b), which has the same mark sequence as the original dataset. The blocks are 
(i) $\{15, 17, 18, 20,\}$ in 2nd track, (ii)
 $\{54, 59, 60\}$ in the 3rd track, (iii)  $\{116, 117, 118, 119, 120, 121, 122, 124\}$ in the 4th track, and (iv) $\{213, 214, 216\}$ in the 5th track.
   By a quick comparisons of  those blocks, one can find that the block $\{15, 17,$ $18, 20\}$ is obtained from the block $\{12, 14, 15, 17\}$ by adding each member by $3$. The second block $\{54, 59, 60\}$ is is obtained by adding 5 into each number in $\{49, 54, 55\}$. In general each block in the transformed dataset in the $t_k^{th}$ track can be obtained by adding a constant $2t_k-1$ into each number in the corresponding block of the original dataset.  
   
   For a translation $T_k$ along $Ray_{i}$ for a given $i<n$, a number $n_j\in X$ with $n_j=q_jn+r_j$, is mapped to $n'_j=q'_jn+r'_j$, where $q'_j = (t_j+k)^2$ and $r'_j = (2(t_j+k)-1)i+(r_j-(2t_j-1)i)$.

  For a scaling $S_k$, a number $n_j\in X$ with $n_j=q_jn+r_j$ is mapped to $n'_j = (t_jk)^2+r_jk$.

\section{Structure of Dataset Connected Components}

 The type-$Q_r$-connected components of a dataset $X$ can be presented as some sort of tree structure. We call it type-II psuedotree, which is slightly different than a pseudotree.

 Given a dataset $X$ on the plane, we can compute the smallest $x$-coordinate and smallest $y$-coordinate by one traversal of $X$. In that way, we can assume that $X$ is in the Quadrant I of an $XY$-coordinate system.
  
   For a real number $r>0$, choose a natural number $n$ such that the length of the line $H^1_0H^1_2$ as in Figure 2(a) is less and equal to $r$. When $n=12$, the circle of radius $r$ is divided into 12 pieces. Then the angle $\angle H^1_0OH^1_2 = 60^\circ$ (See Figure 2(a)) and the triangle $H^1_0OH^1_2$ is an isosceles triangle. Hence the edge has length $H^1_0H^1_2\leq r$. As a result, type-$Q_r$-connected components in track 1 are separated by 4 blank areas with a total distance (i.e., the length of $H^1_0H^1_2$) less than and equal to $r$. Further, a type-$Q_r$-connected component in track 1 is separated from a type $Q_r$-connected component in track 3 by a distance of at least $r$.    
 
 Each type-$Q_r$-connected component in a track $t$ can also be designated by a closed inter interval. Usig the dataset $X$ in Figure 3(a), there is only one type $Q_r$-connected component in track 2, which is $C^2_0=\{S^2_0, S^2_2, S^2_3, S^2_4, S^2_6\}$. As its integer interval, it will be $[12, 17]$. There are two type-$Q_r$-connected components in track 3. They are $C^3_0=\{S^3_1\}, C^3_1=\{S^3_6, S^3_7\}$. They are represented by integer intervals $[49,49]$ and $[54,55]$ respectively. In track 3, there is only one type-$Q_r$-connected component, i.e., $C^4_0=\{S^4_1,S^4_2, S^4_3, S^4_4, S^4_5, S^4_6, S^4_7, S^4_8\}$. Its integer interval is $[193,201]$. The only type-$Q_r$-connected component in track 5 is $C^5_0=\{S^5_{13}, S^5_{14}, S^5_{15}\}$, whose integer interval is $[312,314]$. 
 \begin{figure}
  \includegraphics[width=4in]{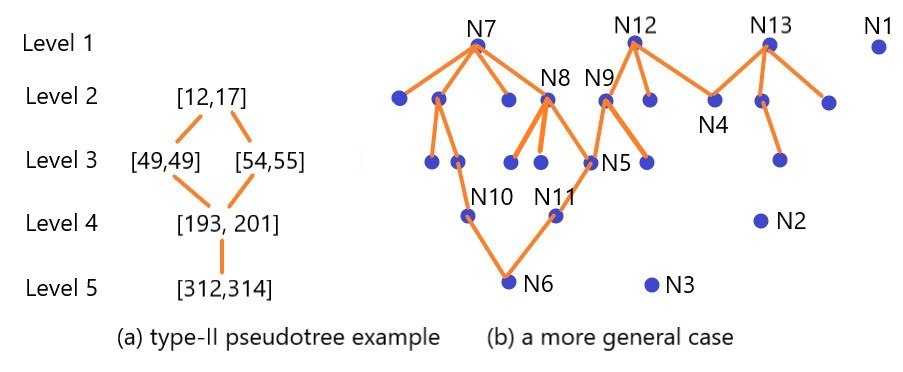}
  \caption{Type-II Pseudotree}
  \label{transformation}
  \end{figure}

 Those type-$Q_r$-connected components can be arranged as a type of tree. The level $t$ of the tree contains the connected components from track $t$. A node $C^t_j$ at level $t$ has a child $C^{t+1}_{j'}$ at level $t+1$ if $C^{t+1}_{j'}\cap (Q_r\vdash CL_1(C^t_j))\neq\emptyset$ (or they are inside a branch). Children nodes of a node are ordered naturally. 
 Two consecutive nodes $C^t_j$ and $C^t_{j+1}$ on the same level $t$ may share at most one child $C^{t+1}_{j'}$. When that happens,  $C^{t+1}_{j'}$ will the last child of $C^t_j$ and the first child of $C^t_{j+1}$. This type of tree is different than the so-called pseudotree. Let's call it type-II pseudotree.
 
 Figure 5(a) shows 
the type-II pseudotree structure of those type-$Q_r$-connected components of $X$ ( Figure 3(a)). Figure 5(b) shows another more general example of a type-II pseudotree. In it, nodes N1, N2, N3 are three anomalies on different tracks. Nodes N8 and N9 share the same child N5. N5 is the last child of N8 and the first child of N9. Nodes N12 and N13 also share the same child N4. The branch from N7 to N10 and then N6 and the branch from N7 to N5, N11 and then N6 form a cycle.

\bibliographystyle{amsplain}

\begin{thebibliography}{99}


\bibitem{carlsson} 
Carlsson, Erik; Carlsson, Gunnar; de Silva, Vin (2006), "An algebraic topological method for feature identification" (PDF), International Journal of Computational Geometry and Applications, 16 (4): 291–314, doi:10.1142/S021819590600204X, S2CID 5831809, archived from the original (PDF) on 2019-03-04.


\bibitem{chambers} 
Chambers, Erin W.; Erickson, Jeff; Worah, Pratik (2008), "Testing contractibility in planar Rips complexes", Proceedings of the 24th Annual ACM Symposium on Computational Geometry, pp. 251–259, CiteSeerX 10.1.1.296.6424, doi:10.1145/1377676.1377721, S2CID 8072058.

\bibitem{chan}
Chan, Timothy M. (1996). "Optimal output-sensitive convex hull algorithms in two and three dimensions". Discrete \& Computational Geometry. 16 (4): 361–368.


\bibitem{chazal}
 Chazal, Frédéric; Oudot, Steve (2008), "Towards persistence-based reconstruction in euclidean spaces", Proceedings of the twenty-fourth annual symposium on Computational geometry, pp. 232–241, arXiv:0712.2638, doi:10.1145/1377676.1377719, ISBN 978-1-60558-071-5, S2CID 1020710.

\bibitem{cover} 
Cover, Thomas M.; Hart, Peter E. (1967). ”Nearest neighbor pattern classification” (PDF).
IEEE Transactions on Information Theory. 13 (1): 21-27.




\bibitem{herbert}
Edelsbrunner, Herbert; Kirkpatrick, David G.; Seidel, Raimund (1983), "On the shape of a set of points in the plane", IEEE Transactions on Information Theory, 29 (4): 551–559,

\bibitem{ester} 
Ester, Martin; Kriegel, Hans-Peter; Sander, Jorg; Xu, Xiaowei (1996). Simoudis, Evangelos;
Han, Jiawei; Fayyad, Usama M. (eds.). A density-based algorithm for discovering clusters
in large spatial databases with noise. Proceedings of the Second International Conference on
Knowledge Discovery and Data Mining (KDD-96). AAAI Press. pp. 226-231, 1996.


\bibitem{freeman}
Freeman, H. (1961). On the encoding of arbitrary geometric configurations. IRE Transactions on Electronic Computers, EC-10(2), 260-268.






\bibitem{graham}
Graham, R.L. (1972). "An Efficient Algorithm for Determining the Convex Hull of a Finite Planar Set" (PDF). Information Processing Letters. 1 (4): 132–133.



  
  \bibitem{hotelling}
 Hotelling, H. (1933). Analysis of a complex of statistical variables into principal components. Journal of Educational Psychology, 24, 417–441, and 498–520.
 
 \bibitem{hotelling2}
Hotelling, H (1936). "Relations between two sets of variates". Biometrika. 28 (3/4): 321–377. doi:10.2307/2333955. JSTOR 2333955.

\bibitem{hu} 
Hu, W., Typed topology and its application on dataset, General Topology and Its Applications, 2024.


\bibitem{jarvis}
Jarvis, R. A.  On the identification of the convex hull of a finite set of points in
the plane. Inform. Process. Lett. 2 (1973), 18–21.





\bibitem{kirkpatrick}
Kirkpatrick, David G.; Seidel, Raimund (1986). "The ultimate planar convex hull algorithm?". SIAM Journal on Computing. 15 (1): 287–299.


\bibitem{kohonen1}
Kohonen, Teuvo (January 2013). "Essentials of the self-organizing map". Neural Networks. 37: 52–65. doi:10.1016/j.neunet.2012.09.018. PMID 23067803. S2CID 17289060.

\bibitem{kohonen2}
Kohonen, Teuvo (2001). Self-organizing maps: with 22 tables. Springer Series in Information Sciences (3 ed.). Berlin Heidelberg: Springer. ISBN 978-3-540-67921-9.







\bibitem{macqueen}
 MacQueen, J. B. (1967). Some Methods for classification and Analysis of Multivariate Observations. Proceedings of 5th Berkeley Symposium on Mathematical Statistics and Probability. Vol. 1. University of California Press. pp. 281–297. MR 0214227. Zbl 0214.46201. Retrieved 2009-04-07.
 


\bibitem{melkemi}
Melkemi M, Djebali M (2000) Computing the shape of a planar points set. Pattern Recogn 33(9):1423–1436


 
\bibitem{pearson}
  Pearson, K. (1901). "On Lines and Planes of Closest Fit to Systems of Points in Space". Philosophical Magazine. 2 (11): 559–572. doi:10.1080/14786440109462720. S2CID 125037489.



\bibitem{rosenfeld}
Rosenfeld, Azriel; Pfaltz, John L. (October 1966). "Sequential Operations in Digital Picture Processing". J. ACM. 13 (4): 471–494.

 
 \bibitem{suzuki}
 Suzuki, S.;  Abe, K. (1985). Topological structural analysis of digitized binary images by border following. Computer Vision, Graphics, and Image Processing, 30(1), 32-46.




\bibitem{zhu1}
Zhu,Jie; Sun, Yizhong; Pang, Yueyong. 2017. A density based algorithm to detect cavities and holes from planar points. Comput. Geosci. 109, C (December 2017), 178–193. https://doi.org/10.1016/j.cageo.2017.08.008






   
   












\end{thebibliography}

\end{document}